\documentclass[runningheads]{llncs}
\usepackage[T1]{fontenc}
\usepackage{graphicx}
\usepackage{orcidlink}
\usepackage{amsmath}
\usepackage{subcaption}
\usepackage{placeins}
\usepackage{float}
\usepackage{tikz}
\usepackage{algorithm}
\usepackage[noend]{algpseudocode} 
\usetikzlibrary{arrows.meta, positioning}

\begin{document}
\title{Fuzzy-RRT for Obstacle Avoidance in a 2-DOF Semi-Autonomous Surgical Robotic Arm}
%
%
\author{
Kaaustaaub Shankar\inst{1}\orcidlink{0009-0000-8970-3746} \and
Wilhelm Louw\inst{1}\orcidlink{0009-0009-8683-1932} \and
Bharadwaj Dogga\inst{1}\orcidlink{0009-0002-5915-1367} \and
Nick Ernest\inst{2}\orcidlink{0000-0002-5819-3541} \and
Tim Arnett\inst{2}\orcidlink{0000-0001-6759-4593} \and
Kelly Cohen\inst{1}\orcidlink{0000-0002-8655-1465}
}

\institute{
College of Engineering and Applied Science, University of Cincinnati, Cincinnati, OH 45219, USA\\
\email{\{shankaks, louwwa, doggabj, cohenky\}@ucmail.uc.edu}
\and
Thales Avionics, Blue Ash, OH 45242\\
\email{\{nick.ernest, tim.arnett\}@defense.us.thalesgroup.com}
}

\authorrunning{K. Shankar et al.}
\maketitle              
\begin{abstract}
AI-driven semi-autonomous robotic surgery is essential for addressing the medical challenges of long-duration interplanetary missions, where limited crew sizes and communication delays restrict traditional surgical approaches. Current robotic surgery systems require full surgeon control, demanding extensive expertise and limiting feasibility in space. We propose a novel adaptation of the Fuzzy Rapidly-exploring Random Tree algorithm for obstacle avoidance and collaborative control in a two-degree-of-freedom robotic arm modeled on the Miniaturized Robotic-Assisted surgical system. It was found that the Fuzzy Rapidly-exploring Random Tree algorithm resulted in an 743 percent improvement to path search time and 43 percent improvement to path cost.
\end{abstract}
\section{Introduction}
The increasing ambition of space agencies and private enterprises to conduct long-duration
interplanetary missions introduces significant medical and surgical challenges. One of the
critical concerns is the need for immediate, lifesaving surgical interventions in environments
where traditional tele-medicine is hindered by substantial communication delays. Given the
constraints of limited crew sizes and the unavailability of multiple medical experts onboard, it is
imperative to develop AI-driven semi-autonomous robotic surgical systems that can operate with
minimal human intervention.
Existing robotic surgical systems are primarily teleoperated, relying on human expertise for
real-time control. However, in microgravity conditions, both the patient and surgical instruments
are subject to unpredictable motion, increasing the risk of errors during procedures. As stated
by Gao et al., the Chief Medical Officer (CMO) of a mission must have an assistant when
treating patients, but the presence of two qualified medical officers cannot be taken for granted
\cite{Gao2017}. To ensure mission success and crew survivability, robotic surgical arms must possess
advanced obstacle avoidance mechanisms, allowing them to navigate complex anatomical
structures, surgical tools, and unforeseen dynamic obstructions with high precision. Current
obstacle avoidance techniques struggle with balancing efficiency and adaptability in highly
constrained surgical environments.
To address this challenge, this research focuses on developing an enhanced path-planning
algorithm for medical robotic systems, particularly for the MIRA Surgical System—an innovative
compact robotic platform designed for remote surgical interventions \cite{virtualincision_mira}. In terms of the level of autonomy, we are aiming for Level 2 where "robots are competent to complete particular surgical activities according to the guidelines given by the physician" \cite{Auto_Rob_Surg_Review}. According to Attanasio et al, "the robot control switches from the human operator to the machine for the duration of the task to be executed" \cite{Attanasio2021}. We see this as the best way to balance human control and automation, allowing both to work together for improved efficiency and safety.

The proposed approach
integrates the Rapidly-exploring Random Tree (RRT) algorithm with fuzzy AI to improve
real-time path planning and obstacle avoidance. By leveraging fuzzy logic, the system can make
adaptive decisions based on environmental uncertainty, leading to smoother, faster, and safer
surgical maneuvers. As demonstrated by Rawat et al. \cite{Rawat_2021}, fuzzy logic significantly enhances the performance of autonomous systems by improving their reliability and practicality. We aim to leverage these benefits to enable advanced surgical capabilities in environments with limited medical resources.

\section{Related Work}

The Rapidly-exploring Random Tree (RRT) algorithm, first introduced by LaValle in 1998 as a
novel approach for path planning \cite{LaValle1998}, has since evolved into a foundational framework for
robotic navigation. Over the years, numerous variants have emerged, such as Informed RRT*
which improved convergence rates through heuristic sampling \cite{Gammell2014}, and RRTX that focused on
real-time replanning capabilities \cite{Otte2015}. Researchers have also integrated additional principles like
attraction and repulsion fields to enhance obstacle avoidance and path optimization \cite{Ma2023};
however, these methods often struggle with dynamic environments due to the need for
continuous recalculations, thus limiting their real-time applicability. This is where fuzzy logic has
shown to be advantageous as it is capable of providing a robust yet efficient solution. Current
implementations of fuzzy in RRT involve other principles such as attraction potential fields as shown in \cite{Wang2024} but
this approach involves applying fuzzy directly to the algorithm without using additional heuristics.

\section{Methodologies}

This paper presents a novel explainable fuzzy-based path planning approach to control a two-degree-of-freedom robotic arm while navigating around obstacles. The methodology integrates the following key components:

\begin{itemize}
    \item \textbf{Rapidly-exploring Random Trees (RRT)}: Utilized to efficiently explore the joint space of the robotic arm.
    \item \textbf{Fuzzy AI}: Enhances the RRT search process by intelligently adjusting parameters for improved performance.
    \item \textbf{Thales True AI}: Optimizes the fuzzy system by determining the best parameters through a Genetic Algorithm (GA).
\end{itemize}

This approach addresses the need for an explainable, low-computation alternative to existing path planning algorithm.

\subsection{Rapidly Exploring Random Trees}
This work employs a standard RRT algorithm for path planning of a two degree-of-freedom
(2-DOF) robotic arm. The joint space is modeled with two axes (x and y) ranging from –180° to
180°. Node expansion is executed by randomly selecting a point within a subset of this joint
space determined by the bounds of the algorithm.  Collision detection is
implemented by verifying whether the end effector of the arm or any of its links come into contact
with obstacles. Because of these steps, a
significant limitation of RRT in real-time applications is its high
computational time which we hope to resolve with fuzzy.

\subsection{Fuzzy AI Enhancement}

To reduce the computation time associated with the standard Rapidly-exploring Random Tree (RRT) algorithm, this work incorporates a fuzzy logic system that adaptively tunes key planning parameters during execution. The fuzzy controller influences three aspects of the search:

\begin{itemize} \item \textbf{Search Bounds}: Continuously adjusts the limits of the search area to prioritize regions likely to contain a valid path. \item \textbf{Goal Bias}: Regulates how often the algorithm samples the goal to encourage faster convergence. \item \textbf{Step Size}: Alters the size of each incremental movement, balancing broad exploration with local precision. \end{itemize}

The fuzzy logic system is implemented using Thales True AI, which trains six parallel Takagi–Sugeno–Kang inference systems. Each system receives four inputs related to the robot's environment:

\begin{itemize} \item The angle and distance to the goal \item The angle and distance to the nearest obstacle \end{itemize}

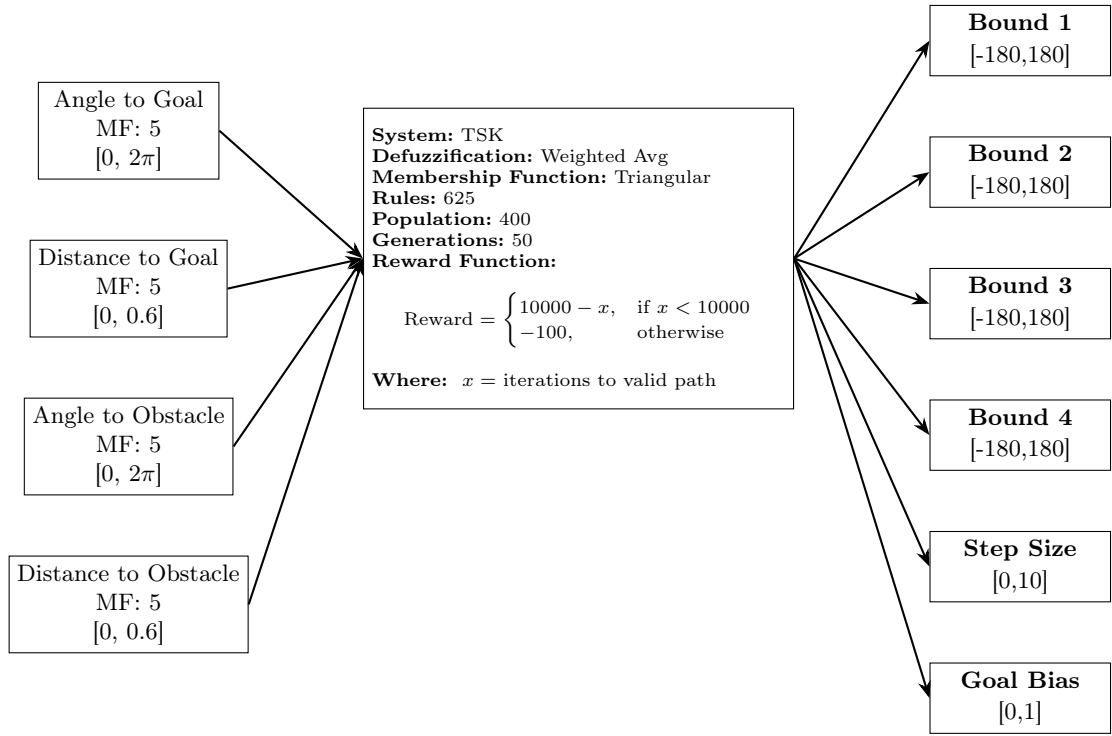
\begin{figure}[htbp]
\centering
\begin{tikzpicture}[
    node distance=0.8cm and 1.8cm,
    box/.style={draw, minimum width=2.4cm, minimum height=0.8cm, font=\small, align=center},
    input/.style={font=\small, align=right},
    arrow/.style={->, thick, >=Stealth}
]

\node[box] (i1) at (0,0) {Angle to Goal\\MF: 5 \\ {[0, \(2\pi\)]}};
\node[box, below=of i1] (i2) {Distance to Goal\\MF: 5 \\ {[0, 0.6]}};
\node[box, below=of i2] (i3) {Angle to Obstacle\\MF: 5 \\ {[0, \(2\pi\)]}};
\node[box, below=of i3] (i4) {Distance to Obstacle\\MF: 5 \\ {[0, 0.6]}};

\node[draw, text width=5.5cm, font=\scriptsize, align=left, minimum height=4cm, right=of i2, yshift=0.4cm] (system) {
    \textbf{System:} TSK\\
    \textbf{Defuzzification:} Weighted Avg\\
    \textbf{Membership Function:} Triangular\\
    \textbf{Rules:} 625\\
    \textbf{Population:} 400\\
    \textbf{Generations:} 50\\
    \textbf{Reward Function:} 
    \[
    \text{Reward} = 
    \begin{cases}
    10000 - x, & \text{if } x < 10000 \\
    -100, & \text{otherwise}
    \end{cases}
    \]
    \textbf{Where: } \(x = \) iterations to valid path
};

\foreach \x in {i1,i2,i3,i4} {
    \draw[arrow] (\x.east) -- (system.west);
}

\node[box, right=of system, yshift=2.9cm] (o1) {\textbf{Bound 1}\\[1pt] [-180,180]};
\node[box, below=of o1] (o2) {\textbf{Bound 2}\\[1pt] [-180,180]};
\node[box, below=of o2] (o3) {\textbf{Bound 3}\\[1pt] [-180,180]};
\node[box, below=of o3] (o4) {\textbf{Bound 4}\\[1pt] [-180,180]};
\node[box, below=of o4] (o5) {\textbf{Step Size}\\[1pt] [0,10]};
\node[box, below=of o5] (o6) {\textbf{Goal Bias}\\[1pt] [0,1]};

\foreach \out in {o1,o2,o3,o4,o5,o6} {
    \draw[arrow] (system.east) -- (\out.west);
}

\end{tikzpicture}
\caption{Fuzzy Inference System for Parameter Generation}
\end{figure}

These FISs are then used to control the creation of the tree in the fuzzy RRT framework, where each FIS influences key sampling and steering behaviors as shown in Algorithm~\ref{alg:fuzzy_rrt}.
\begin{algorithm}[H]
\caption{Fuzzy RRT Tree Construction: \texttt{build\_tree(F)}}
\label{alg:fuzzy_rrt}
\begin{algorithmic}[1]
\State $nearest\_node \gets \textit{start}$

\For{$iteration = 1$ to $max\_iter$}
    \State $ee\_pos \gets$ \Call{ForwardKinematics}{$nearest\_node$}
    
    \If{obstacles exist}
        \State $closest\_obstacle \gets$ nearest obstacle to $ee\_pos$
        \State $angle\_to\_obs \gets$ angle from $ee\_pos$ to $closest\_obstacle$
        \State $dist\_to\_obs \gets$ distance from $ee\_pos$ to $closest\_obstacle$
    \Else
        \State $angle\_to\_obs \gets 0$
        \State $dist\_to\_obs \gets \infty$
    \EndIf

    \State $goal\_pos \gets$ \Call{ForwardKinematics}{$goal$}
    \State $angle\_to\_goal \gets$ angle from $ee\_pos$ to $goal\_pos$
    \State $dist\_to\_goal \gets$ distance from $ee\_pos$ to $goal\_pos$

    \State $(bias, B_1, B_2, B_3, B_4, step) \gets$ \Call{F}{$angle\_to\_obs$, $dist\_to\_obs$, $angle\_to\_goal$, $dist\_to\_goal$}
    \State $bounds \gets \big[[B_1, B_2], [B_3, B_4]\big]$

    \State $sample \gets$ \Call{SamplePoint}{$bias$, $bounds$}
    \State $nearest\_node \gets$ \Call{NearestNeighbor}{$sample$}
    \State $new\_node \gets$ \Call{Steer}{$nearest\_node$, $sample$}

    \If{\textbf{not} \Call{IsCollisionFree}{$new\_node$}} \Comment{Node in collision}
        \State \textbf{continue}
    \EndIf

    \If{\textbf{not} \Call{IsPathCollisionFree}{$nearest\_node$, $new\_node$}} \Comment{Path blocked}
        \State \textbf{continue}
    \EndIf

    \State Add $new\_node$ to tree
    \State $parent[new\_node] \gets nearest\_node$

    \If{$new\_node == goal$}
        \State \Return success
    \EndIf

    \If{\Call{Distance}{$new\_node$, $goal$} $\leq step$}
        \If{\Call{IsPathCollisionFree}{$new\_node$, $goal$}}
            \State Add $goal$ to tree
            \State $parent[goal] \gets new\_node$
            \State \Return success
        \EndIf
    \EndIf
\EndFor

\State \Return failure
\end{algorithmic}
\end{algorithm}

\subsection{Genetic Algorithm (GA)}

The Thales True AI framework optimizes fuzzy systems using a Genetic Algorithm (GA), which was employed to train a population of 400 over 50 generations or until a fitness of 9750 was achieved. This threshold was selected as a practical stopping criterion, as each generation required approximately five hours to complete, making a full 50-generation run time-prohibitive. Additionally, limiting the number of generations helped mitigate the risk of overfitting. While the specific fitness score of 9750 was somewhat arbitrary, it represented a reasonable compromise between achieving acceptable performance and operating within time and computational constraints. With five triangular membership functions per FIS—where the leftmost and rightmost serve as shoulder functions—we train a total of 637 parameters per FIS. This includes 3 membership function parameters and $5^4$ fuzzy rules. Given that the system employs six FISs, one for each output parameter, this results in a total of 3,822 genes per individual in the evolutionary population. To ensure consistency despite the inherent randomness of RRT, the GA's fitness function is the average reward over five simulation runs. This prevents less fit chromosomes from being favored due to chance. The reward for each simulation is calculated using the relation:

\begin{equation}
\text{Reward} = \begin{cases}
10000 - \text{x}, & \text{if x} < 10000 \\
-100, & \text{otherwise}
\end{cases}
\end{equation}

where x represents the number of iterations needed to find a valid path. This formulation ensures that any path found is preferable to no path at all. By maximizing this function, the GA selects solutions that reduce the average number of iterations, improving computational efficiency.

For simulation and evaluation, the path planning and Fuzzy AI integration are implemented in Python. Key performance metrics include the minimum, maximum, and median iteration count of the algorithm over 50 runs. These metrics benchmark the improvements introduced by the fuzzy components compared to the traditional RRT method.

\FloatBarrier
\section{Results}
\begin{figure}[htbp]
    \centering
    \includegraphics[width=0.8\textwidth]{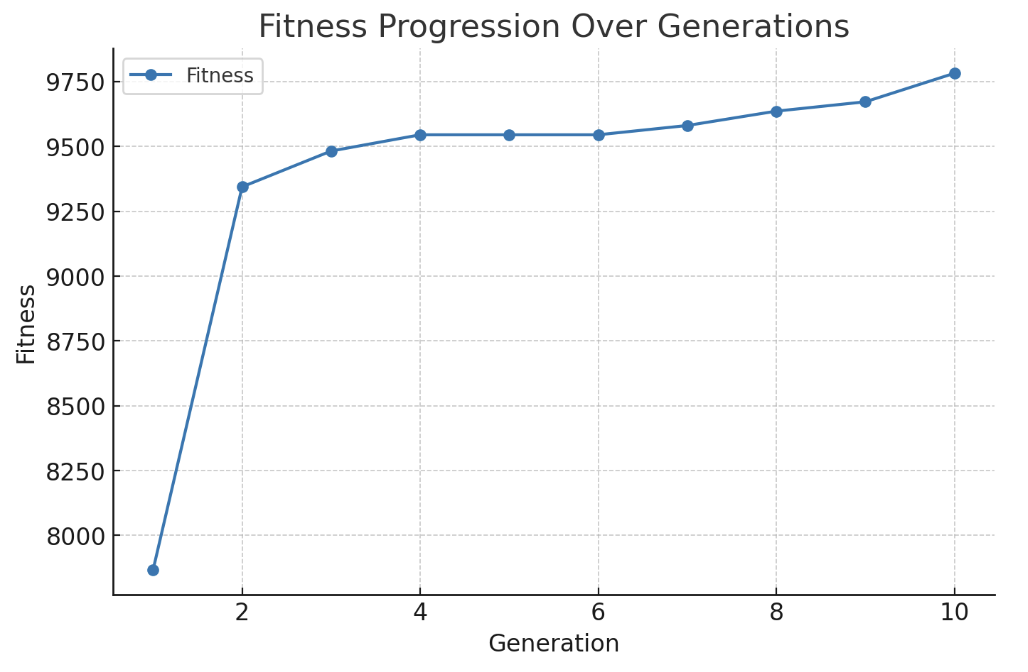}
    \caption{Fitness of GFS over 10 Generations }
    \label{fig:normal}
\end{figure}

\begin{figure}[H]
    \centering
    \begin{subfigure}[b]{0.48\textwidth}
        \centering
        \includegraphics[width=\textwidth]{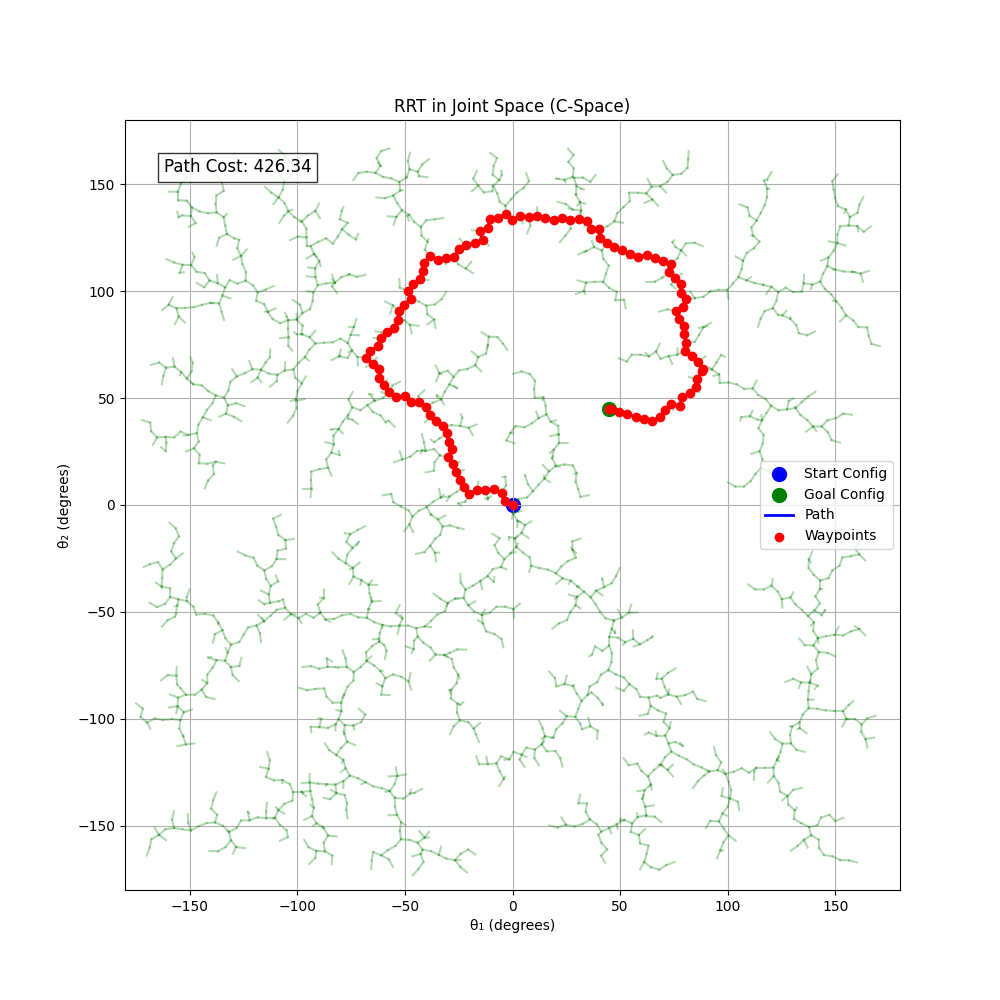}
        \caption{Path generated using RRT.}
        \label{fig:rrt_path}
    \end{subfigure}
    \hfill
    \begin{subfigure}[b]{0.48\textwidth}
        \centering
        \includegraphics[width=\textwidth]{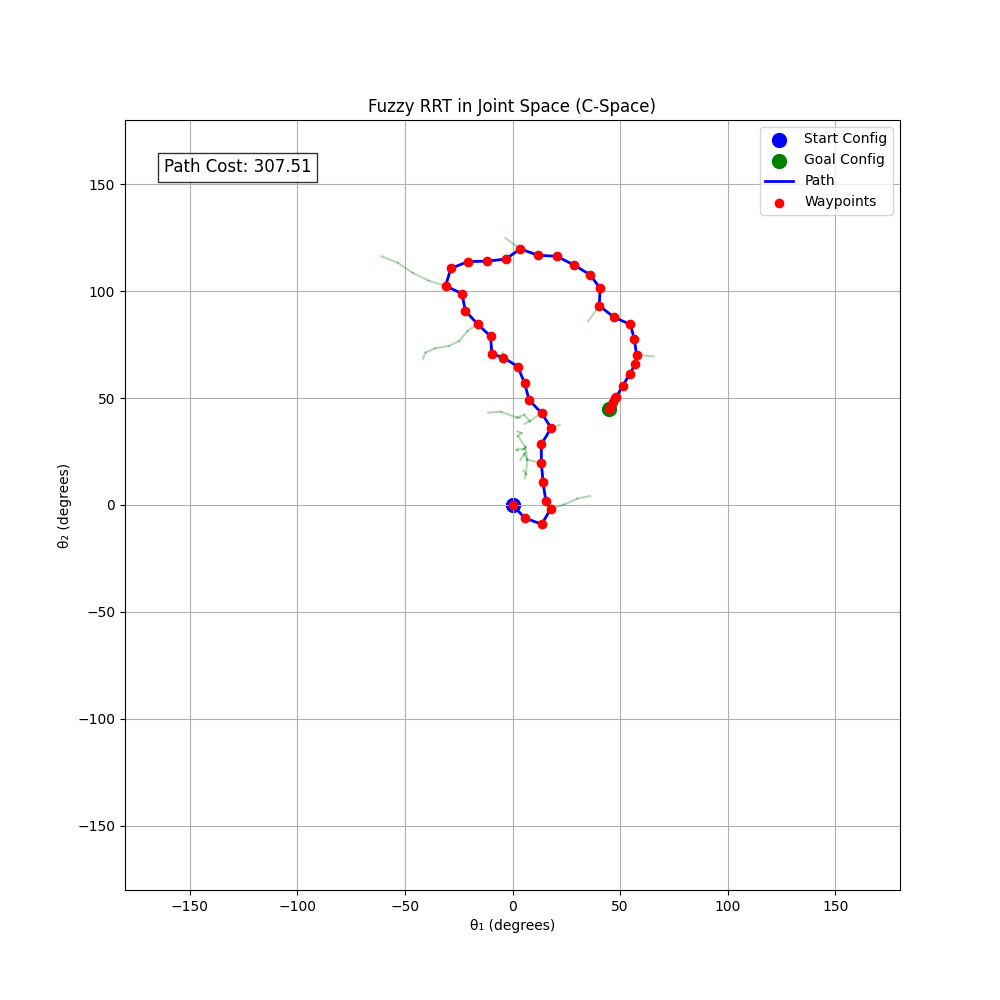}
        \caption{Path generated using Fuzzy RRT.}
        \label{fig:fuzzy_rrt_path}
    \end{subfigure}
    \caption{Comparison of path planning using standard RRT vs. Fuzzy RRT.}
    \label{fig:rrt_comparison}
\end{figure}
\vspace{-1.5em}
\begin{figure}[hbp]
    \centering
    \begin{subfigure}[b]{0.48\textwidth}
        \centering
        \includegraphics[width=\textwidth]{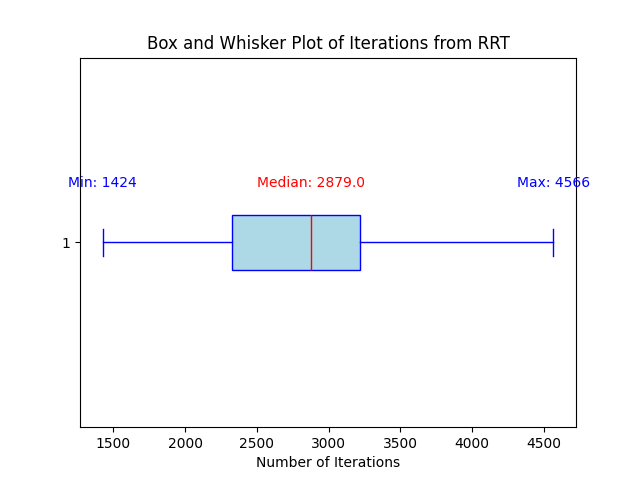}
        \caption{Performance metrics of RRT.}
        \label{fig:rrt_boxplot}
    \end{subfigure}
    \hfill
    \begin{subfigure}[b]{0.48\textwidth}
        \centering
        \includegraphics[width=\textwidth]{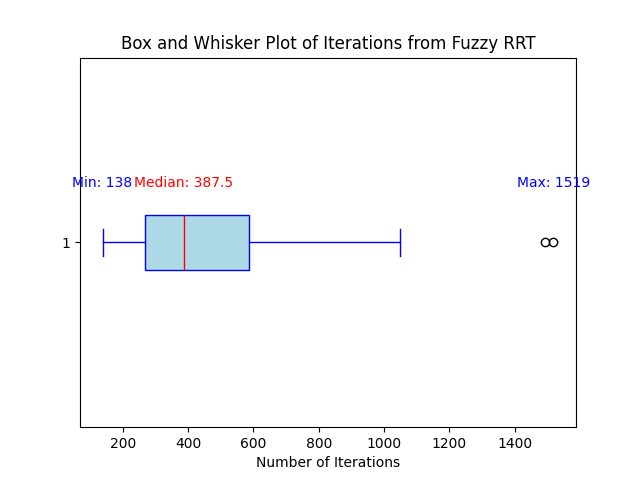}
        \caption{Performance metrics of Fuzzy RRT.}
        \label{fig:fuzzy_boxplot}
    \end{subfigure}
    \caption{Comparison of Iteration metrics between standard RRT and Fuzzy RRT.}
    \label{fig:performance_comparison}
\end{figure}

\begin{figure}[H]
    \centering
    \begin{subfigure}[b]{0.48\textwidth}
        \centering
        \includegraphics[width=\textwidth]{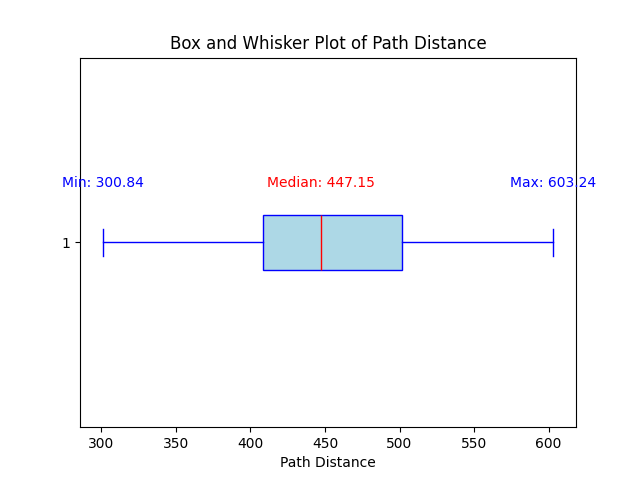}
        \caption{Performance metrics of RRT.}
        \label{fig:rrt_boxplot}
    \end{subfigure}
    \hfill
    \begin{subfigure}[b]{0.48\textwidth}
        \centering
        \includegraphics[width=\textwidth]{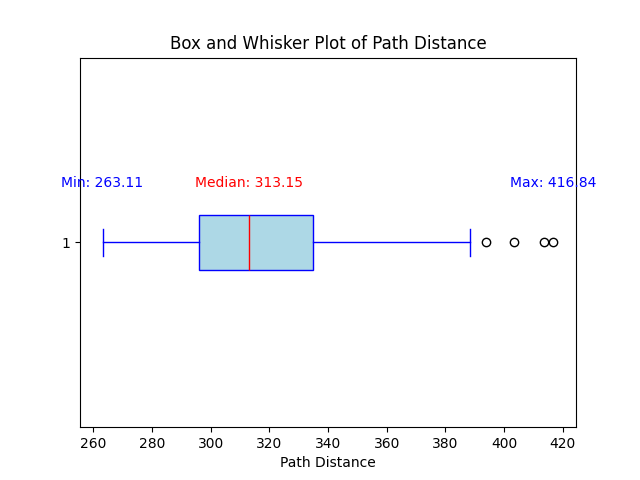}
        \caption{Performance metrics of Fuzzy RRT.}
        \label{fig:fuzzy_boxplot}
    \end{subfigure}
    \caption{Comparison of Path Cost between standard RRT and Fuzzy RRT.}
    \label{fig:cost_comparison}
\end{figure}

As can be seen in Figure~\ref{fig:normal}, the 4 in 6 out system performs well and easily achieved the desired
fitness of 9750 in 10 generations and we are able to see the performance difference through Figure~\ref{fig:rrt_comparison}. Not only was the algorithm able to find a path with a lower cost, it was able to do so
without searching as much which is clearly shown in the green. After testing the algorithm on 200
runs and collecting the results in Figure~\ref{fig:performance_comparison} and Figure~\ref{fig:cost_comparison}, we can see nearly a 8-fold reduction in iterations needed
indicating the value of having fuzzy control the algorithm.
Compared with the results of Wang et al. where they were able to reduce 'the average initial path search
time by approximately 46.52\% and the average final path cost by approximately 10.01\%', we have been able to improve the search time by 743\% and the cost of the path by 43\% \cite{Wang2024}. 
These results make it clear: Fuzzy RRT isn’t just better—it’s the new standard for fast, high-quality pathfinding.

\section{Future Work}
Since the original project centers on the MIRA robotic arm, the next step is to expand the algorithm to handle five degrees of freedom (5-DOF) while integrating a second, identical arm for collaborative robotics. To achieve this, the new system will be trained across a wide range of simulated scenarios, allowing the fuzzy inference system (FIS) to generalize across complex and dynamic environments. Additionally, we can implement a form of informed RRT where a fuzzy-enhanced RRT is used to quickly generate a feasible path toward the goal. Once a path is found, the fuzzy system can then refine it using an optimization step, smoothing and adjusting the trajectory based on a predefined cost function—such as minimizing joint effort, maximizing clearance, or balancing task load between the two arms.  Furthermore, we will be testing the system’s robustness to noise and dynamic obstacles through Stochastic Robustness Analysis.

\begin{credits}
\subsubsection{\ackname}

The authors extend their sincere gratitude to the members of the AI Bio Lab at the University of Cincinnati for their invaluable discussions and collaborative efforts that facilitated the realization of this work. In particular, the contributions of Magnus Sieverding, Jared Burton, Lohith Pentapalli, Tri Nguyen, Garrett Olges, Nate Steffan, Lucia Vilar Nuño, and Hugo Henry are highly appreciated.

The authors express their appreciation to Thales for providing licenses to utilize its TRUE AI framework. Without this assistance, the development of this controller would have been significantly more arduous and the outcomes would have been substantially diminished.

We acknowledge the support provided by the ASTRO Discover Fellowship (Armstrong Institute for Space, Research, and Technology), which has provided the requisite resources and funding to enable this research. We would like to express our specific gratitude to Professor Charles R Doarn.

\subsubsection{\discintname}
The authors have no competing interests to disclose.
\end{credits}
%
%
%
%

\bibliographystyle{unsrt}
\bibliography{Bib}

\begin{thebibliography}{10}

\bibitem{Gao2017}
Y.~Gao and S.~Chien.
\newblock Review on space robotics: Toward top-level science through space exploration.
\newblock {\em Science Robotics}, 2(7), 2017.

\bibitem{virtualincision_mira}
{Virtual Incision Corporation}.
\newblock Mira surgical robot.
\newblock \url{http://virtualincision.com/mira/}, 2025.
\newblock Accessed: 2025-04-08.

\bibitem{Auto_Rob_Surg_Review}
Yeisson Rivero-Moreno, Miguel Rodriguez, Paola Losada-Muñoz, Samantha Redden, Saiddys Lopez-Lezama, Andrea Vidal-Gallardo, Debbye Machado-Paled, Jesus Cordova~Guilarte, and Sheyla Teran-Quintero.
\newblock Autonomous robotic surgery: Has the future arrived?
\newblock {\em Curēus (Palo Alto, CA)}, 16(1):e52243--e52243, 2024.

\bibitem{Attanasio2021}
A.~Attanasio, B.~Scaglioni, E.~De~Momi, P.~Fiorini, and P.~Valdastri.
\newblock Autonomy in surgical robotics.
\newblock {\em Annual Review of Control, Robotics, and Autonomous Systems}, 4(1):651--679, 2021.

\bibitem{Rawat_2021}
Devendra Rawat, Mukul~K. Gupta, and Abhinav Sharma.
\newblock Intelligent control of robotic manipulators: a comprehensive review.
\newblock {\em Spatial information research (Online)}, 31(3):345--357, 2023.

\bibitem{LaValle1998}
S.~LaValle.
\newblock Rapidly-exploring random trees: A new tool for path planning.
\newblock {\em Technical Report, University of Illinois}, 1998.

\bibitem{Gammell2014}
J.~D. Gammell, S.~S. Srinivasa, and T.~D. Barfoot.
\newblock Informed rrt*: Optimal sampling-based path planning focused via direct sampling of an admissible ellipsoidal heuristic.
\newblock {\em IEEE/RSJ International Conference on Intelligent Robots and Systems}, pages 2997--3004, 2014.

\bibitem{Otte2015}
M.~Otte and E.~Frazzoli.
\newblock Rrtx real-time motion planning/replanning for environments with unpredictable obstacles.
\newblock {\em Springer Tracts in Advanced Robotics}, pages 461--478, 2015.

\bibitem{Ma2023}
B.~Ma, C.~Wei, Q.~Huang, and J.~Hu.
\newblock Apf-rrt*: An efficient sampling-based path planning method with the guidance of artificial potential field.
\newblock {\em International Conference on Mechatronics and Robotics Engineering (ICMRE)}, pages 207--213, 2023.

\bibitem{Wang2024}
H.~Wang, X.~Zhou, J.~Li, Z.~Yang, and L.~Cao.
\newblock Improved rrt* algorithm for disinfecting robot path planning.
\newblock {\em Sensors}, 24(5):1520, 2024.

\end{thebibliography}

\end{document}